\documentclass[11pt]{article}
\usepackage{arabtex}
\usepackage{acolor} 
\usepackage[]{acl}

\usepackage{tipa}

\usepackage{times}
\usepackage{latexsym}

\usepackage[T1]{fontenc}
\usepackage[utf8]{inputenc}
\usepackage{microtype}

\usepackage{times}
\usepackage{multirow}
\usepackage{amsmath}
\usepackage{amssymb}
\usepackage{subcaption}
\usepackage{tikz}
\usepackage{pgfplots}
\usepackage{latexsym}

\usepackage{etoolbox}
\usepackage{todonotes}
\usepackage{microtype}
\usepackage{comment}
\usepackage{wasysym}

\newcommand{\ipa}{\textipa} 

 \makeatletter
 \newcommand{\printfnsymbol}[1]{%
 	\textsuperscript{\@fnsymbol{#1}}%
 }
\makeatother

\title{Automatic Standardization of Colloquial Persian}

\author{Mohammad Sadegh Rasooli$^{1}$\ \ 
Farzane Bakhtyari$^2$\thanks{~~Equal contribution in data annotation.} \ \ Fatemeh Shafiei$^{3}$\printfnsymbol{1}   \\
\textbf{Mahsa Ravanbakhsh$^{3}$\printfnsymbol{1} \and Chris Callison-Burch$^{1}$}\\
	\normalsize{~$^1$~Department of Computer and Information Science,  University of Pennsylvania, Philadelphia, PA, USA}\\[-.05cm]
	\normalsize{~$^2$~Institute for Humanities and Cultural Studies, Tehran, Iran}\\[-.05cm]
    \normalsize{~$^3$~Institute for Cognitive Science Studies, Tehran, Iran}\\[-.05cm]
	{\small\tt \{rasooli, ccb\}@seas.upenn.edu,\  f.bakhtyari@ihcs.ac.ir}  \\[-.09cm] {\small\tt  \{shafiei\_f, ravanbakhsh\_m\}@iricss.org }
}
\date{}

\begin{document}
\maketitle
\begin{abstract}
The Iranian Persian language has two varieties: standard and colloquial. Most natural language processing tools for Persian assume that the text is in standard form: this assumption is wrong in many real applications especially web content. This paper describes a simple and effective standardization approach based on sequence-to-sequence translation. We design an algorithm for generating artificial parallel colloquial-to-standard data for learning a sequence-to-sequence model. Moreover, we annotate a publicly available evaluation data consisting of 1912 sentences from a diverse set of domains. Our intrinsic evaluation shows a higher BLEU score of 62.8 versus 61.7 compared to an off-the-shelf rule-based standardization model in which the original text has a BLEU score of 46.4. We also show that our model improves English-to-Persian machine translation in scenarios for which the training data is from colloquial Persian with 1.4 absolute BLEU score difference in the development data, and 0.8 in the test data.
\end{abstract}

\section{Introduction}\label{sec:intro}
There has recently been a great deal of interest in developing natural language processing (NLP) datasets and models for the Persian language (e.g. \cite{bijankhan2011lessons,rasooli2013development,feely2014cmu,seraji2015morphosyntactic,nourian2015importance,seraji2016universal,mirzaei2016persian,mirzaei2018persian,poostchi2018bilstm,mirzaei2020semantic,taher2020beheshti,khashabi20}). Despite impressive achievements, most of Persian NLP models assume that the text is written in the \emph{standard form}. This assumption is practically not true~\cite{solhju}. This particular phenomenon is somewhat similar to dialectal variations in Arabic, but it is less diverse than what we observe in Arabic dialects. The general upshot is that many natural language processing systems for Persian might easily break when dealing with a mixture of colloquial and standard text. This is indeed an important issue given the surge in using social media and online contents.

The focus of this paper, similar to the majority of previous work, is on contemporary Iranian Persian. Persian is an Indo-Eurpoean language with more than 100 million speakers across the world especially in Iran, Afghanistan, and Tajikistan~\cite{windfuhr2009iranian}.  Loosely speaking, Iranian Persian is used in two different but similar forms: \emph{standard} and \emph{colloquial}~\cite{boyle1952notes,shamsfard2011challenges,tabib_book}. In some sense, this idiosyncratic phenomenon is a type of \emph{diglossia} for which there exists a \emph{high variety} of language used in formal writing, and a \emph{low variety} used in spoken language~\cite{jeremias1984diglossia,bakh2018}. What we refer here as \emph{colloquial Persian} is a version of the spoken language that is mostly used in Tehran (captial of Iran). However, due to the national academic system and media, most people in Iran understand and use it in daily basis~\cite{solhju}. 

Colloquial Persian has many broken words for which some of them invent their own morphology, and sometimes idiosyncratic syntactic order, and some new idioms~\cite{tabib_book}. It is diverse and appears in different shapes from merely standard syntax with some occasional broken words (as used in TV and radio news) to broken words with colloquial syntax. Depending on the formality of context and personal writing style, a colloquial sentence might have some few broken words and few other standard word forms: This is somewhat like code-switching between two varieties of the same language instead of two distinct languages.

This paper proposes an automatic method for standardizing colloquial Persian text. The core idea behind our work is training a sequence-to-sequence translation model~\cite{vaswani2017attention} that translates colloquial Persian to standard Persian. We believe that our approach is a more viable technique than merely applying some conversion rules for standardization. On the one hand, writing an extensive set of rules depending on different semantic contexts is cumbersome, if not impossible. On the other hand, using a sequence-to-sequence model facilitates leveraging pretrained language models such as masked language models~\cite{devlin-etal-2019-bert} trained on huge amount of monolingual texts.  Our experiments also proves this point. Since there is no available parallel text between standard and colloquial Persian, we propose a random standard-to-colloquial conversion algorithm based on recent linguistic studies on common colloquial word forms in Persian literature~\cite{tabib_journal,tabib_book}.  

Our contribution is three-fold: 1) We propose a novel method for training translation models from colloquial to standard Persian, and show improvements over an off-the-shelf rule-based model. Except a paper written in Persian with a simple n-gram approach~\cite{armin} and without any code or data release, we are not aware of any work on this topic; 2) We provide a publicly available manually annotated evaluation data for colloquial Persian with two types of standardization for which the first only concerns surface word forms, and the second considers making the text stylistically standard. This data consists of 1929 sentences from different genres including fiction, translated fiction, blog posts, public group chats, and comments in news websites;\footnote{Available for download from \url{https://github.com/rasoolims/Shekasteh}} and  3) We show that our method improves English-to-Persian machine translation trained on colloquial Persian data. 
\section{Automatic Standardization and Evaluation}\label{sec:approach}
There are three core components of our research: 1) Modeling, 2) Training data, 3) Evaluation. This section briefly describes these three components.

\subsection{Model}
We model colloquial-to-standard text conversion as machine translation. The input to the system is a sequence of colloquial words $\{x_1\ldots x_n\}$ and the output is a sequence of standard word forms $\{y_1\ldots y_m\}$ where $n$ and $m$ are not necessarily equal. We train a standard neural machine translation model with attention~\cite{vaswani2017attention} on a training data in which each colloquial sentence is accompanied by its standard version. Figure~\ref{fig:break_example} shows examples of such training data. \emph{Neural machine translation} uses sequence-to-sequence models with attention~\cite{cho-etal-2014-learning,Bahdanau2015NeuralMT,vaswani2017attention} for which the likelihood of training data ${\cal D} = \{(x^{(1)}, y^{(1)}) \ldots (x^{(|{\cal D}|)}, y^{(|{\cal D}|)})\}$ is maximized by maximizing the log-likelihood of predicting each target word given its previous predicted words and source sequence:
\[
{\cal L}({\cal D}) = \sum_{i=1}^{|{\cal D}|} \sum_{j=1}^{|y^{(i)}|} \log p(y^{(i)}_{j} | y^{(i)}_{k<j}, x^{(i)}; \theta)
\]
where $\theta$ is a collection of parameters to be learned.

\subsection{Training Data Generation}

 Since there is no parallel data to train a machine translation model, we define a set of rules to break standard forms into colloquial forms. In general, there are more than 300 rules, mostly inspired from \newcite{tabib_journal}. Table~\ref{tab:rules} lists the main rules used in this work.\footnote{The code for this rules is publicly available in \url{https://github.com/rasoolims/PBreak}.} These rules include changing vowels (e.g. \ipa{a}$\rightarrow$\ipa{u}), modifying verb forms (e.g. \setfarsi\novocalize \<bgwym> [\ipa{beg\textcolor{red!60!}{uy}\ae m}] $\rightarrow$ \setfarsi\novocalize \<bgm> [\ipa{beg\ae m}]), and other types of conversion listed by \newcite{tabib_journal}. 

\begin{figure*}
    \centering
    \includegraphics[width=1.01\textwidth]{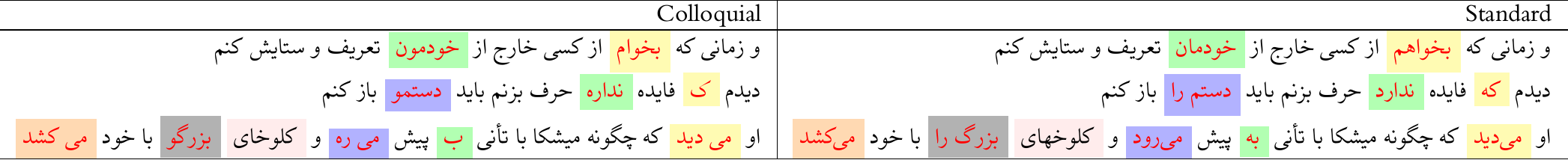}
    \caption{Real examples of our random conversions of standard text to colloquial.}
    \label{fig:break_example}
\end{figure*}

Our conversion rules define a function $f(x_{i}, x_{i+1})$ in which depending on part-of-speech tag and the next word $x_{i+1}$ form, a new word form $y$ or a sequence of word forms are generated. Since we are not sure about the extent in which words are broken in text, we randomly skip some conversions with a probability of $0.1$ in order to make the text look like a mix of colloquial and standard word forms. One advantage of this approach is that we can easily have a large amount of parallel text to train a neural machine translation model. Some real examples are shown in Figure~\ref{fig:break_example}. There are definitely many cases for which our conversion rules create a wrong colloquial word form. However, since the standard form is the ultimate output that we care about, there is less risk of making conversion mistakes. This is very similar to back-translation~\cite{sennrich-etal-2016-improving} for which the input of the model is a machine translated text with potential errors.

\begin{table}[t!]
\renewcommand{\arraystretch}{1.3} 
    \centering \small 
      \setlength{\tabcolsep}{0pt}
    \begin{tabular}{l|c c}
 \hline \hline
    \multirow{2}{*}{Rule name} & \multicolumn{2}{c}{Example} \\ 
     &  Standard &  Colloquial \\ \hline
    \multirow{3}{*}{ ``\textcolor{red}{\ipa{an}}'' suffix}   & \setfarsi\novocalize \<thrAn> [\ipa{tehr\textcolor{red}{a}n}] & \setfarsi\novocalize \<thrwn> [\ipa{tehr\textcolor{red}{u}n}]  \\
    & \setfarsi\novocalize \<n^sAn> [\ipa{neS\textcolor{red}{a}n}] & \setfarsi\novocalize \<n^swn> [\ipa{neS\textcolor{red}{u}n}]  \\
    &  \setfarsi\novocalize \<gmAn> [\ipa{g\ae m\textcolor{red}{a}n}] & \setfarsi\novocalize \<gmwn> [\ipa{g\ae m\textcolor{red}{u}n}]  \\ \hline
    \multirow{3}{*}{ Verb suffix} & \setfarsi\novocalize \<brwm> [\ipa{ber\ae\textcolor{red}{v\ae}m}]  & \setfarsi\novocalize \<brm> [\ipa{ber\ae m}] \\
    &  \setfarsi\novocalize \<brwd> [\ipa{ber\textcolor{red}{\ae v\ae d}}]  & \setfarsi\novocalize \<brh> [\ipa{ber\textcolor{red}{e}}]\\
       &  \setfarsi\novocalize \<brwnd> [\ipa{ber\ae\textcolor{red}{v\ae nd}}]  & \setfarsi\novocalize \<brn> [\ipa{ber\ae\textcolor{red}{n}}] \\ \hline
    \multirow{4}{*}{Verb form} & \setfarsi\novocalize \<mygwym> [\ipa{mig\textcolor{red}{uy}\ae m}] & \setfarsi\novocalize \<mygm> [\ipa{mig\ae m}]   \\
& \setfarsi\novocalize \<.AystAdm> [\ipa{\textcolor{red}{Pist}ad\ae m}] & \setfarsi\novocalize \<wAysAdm> [\ipa{\textcolor{red}{vays}ad\ae m}]   \\
&  \setfarsi\novocalize \<bxwAhm> [\ipa{bexa\textcolor{red}{h\ae}m}] & \setfarsi\novocalize \<bxAm> [\ipa{bexam}]  \\
&  \setfarsi\novocalize \<Amdm> [\ipa{\textcolor{red}{Pa}m\ae d\ae m}] & \setfarsi\novocalize \<.Awmdm> [\ipa{\textcolor{red}{Pu}m\ae d\ae m}]  \\  \hline
     \multirow{2}{*}{``\textcolor{red}{\ipa{ha}}'' suffix }  & \setfarsi\novocalize \<glhA> [\ipa{gol\textcolor{red}{ha}}] & \setfarsi\novocalize \<glA> [\ipa{gol\textcolor{red}{a}}]   \\  
    &  \setfarsi\novocalize \<glhAy^y> [\ipa{gol\textcolor{red}{ha}i}] & \setfarsi\novocalize \<glAy^y> [\ipa{gol\textcolor{red}{a}i}]   \\ \hline
\multirow{4}{*}{Common rules} & \setfarsi\novocalize \<.sA.hb> [\ipa{saheb}] & \setfarsi\novocalize \<.sA.hAb> [\ipa{sah\textcolor{red}{a}b}]   \\
& \setfarsi\novocalize \<dygr> [\ipa{dig\textcolor{red}{\ae r}}] & \setfarsi\novocalize \<dygh> [\ipa{dig\textcolor{red}{e}}]   \\
&  \setfarsi\novocalize \<AnjA> [\ipa{\textcolor{red}{Pa}nja}] & \setfarsi\novocalize \<awnjA> [\ipa{\textcolor{red}{Pu}nja}]  \\
&  \setfarsi\novocalize \<rA> [\ipa{r\textcolor{red}{a}}] & \setfarsi\novocalize \<rw> [r\ipa{\textcolor{red}{o}}]  \\  \hline
 \multirow{2}{*}{Case marker}    &   \setfarsi\novocalize \<tw rA> [\ipa{to r\textcolor{red}{a}}] & \setfarsi\novocalize \<twrw> [tor\ipa{\textcolor{red}{o}}]  \\  
  &  \setfarsi\novocalize \<An rA> [\ipa{Pan \textcolor{red}{ra}}] & \setfarsi\novocalize \<.Awnw> [\ipa{P\textcolor{red}{uno}}]   \\ \hline
\multirow{2}{*}{Attach pronoun}     &   \setfarsi\novocalize \<bh tw> [\ipa{be   \textcolor{red}{tu}}] & \setfarsi\novocalize \<bht> [\ipa{be\textcolor{red}{het}}]  \\
&    \setfarsi\novocalize \<bh .Aw> [\ipa{be   \textcolor{red}{Pu}}] & \setfarsi\novocalize \<bh^s> [\ipa{be\textcolor{red}{heS}}]  \\ \hline
 ``\textcolor{red}{\ipa{P\ae st}}'' (is)  &   \setfarsi\novocalize \<km .Ast> [\ipa{k\ae m   \textcolor{red}{P\ae st}}] & \setfarsi\novocalize \<kmh> [\ipa{k\ae m\textcolor{red}{e}}]  \\ \hline
 \multirow{2}{*}{``\textcolor{red}{\ipa{h\ae st}}'' (is) }&   \setfarsi\novocalize \<km hstnd> [\ipa{k\ae m   \textcolor{red}{h\ae st\ae nd}}] & \setfarsi\novocalize \<kmnd> [\ipa{k\ae m\textcolor{red}{\ae nd}}]  \\
 & \setfarsi\novocalize \<km hsty> [\ipa{k\ae m   \textcolor{red}{h\ae sti}}] & \setfarsi\novocalize \<kmy> [\ipa{k\ae m\textcolor{red}{i}}]  \\ \hline \hline

    \end{tabular}
    \caption{Some of the main conversion rules for converting standard Persian to colloquial with examples. Most of these rules are inspired from \protect\newcite{tabib_journal}.}
    \label{tab:rules}
\end{table}

\subsection{Evaluation Data}
We collect data from different sources including comments on news websites, discussions on web forums, translated fiction, and dialogues in Persian fiction. Three native speaker linguists have annotated the data in which the development part (917 sentences) is annotated by the first annotator, and the test data (1012 sentences) is annotated by the other two annotators. For the sake of keeping sentences coherent, many sentences are actually multiple short sentences. Figure~\ref{fig:genre} shows the counts of sentences in different topics. As seen in the Figure, the distributions of the development and test datasets are intentionally very different.

\usetikzlibrary{patterns}
\begin{figure}
    \centering

\scalebox{0.5}{
\pgfplotstableread[row sep=\\,col sep=&]{
    direction & Development Data & Test Data \\
   \rotatebox{90}{Telegram groups} & 0 & 100 \\
    \rotatebox{90}{Movie subtitles} & 52 & 101 \\  
     \rotatebox{90}{Translated fiction} & 0 & 94  \\ 
     \rotatebox{90}{Persian fiction} & 0 & 152   \\ 
     \rotatebox{90}{Blog posts} & 0 & 180 \\ 
     \rotatebox{90}{News comments} & 33 & 385   \\ 
\rotatebox{90}{Product reviews} & 41 & 0   \\ 
\rotatebox{90}{Online discussions} & 791 & 0    \\ 
}\mydata

\begin{tikzpicture}
    \begin{axis}[
            ybar,
            bar width=.6cm,
            width=\textwidth,
            height=.5\textwidth,
            legend style={at={(0.65,0.95)},
                anchor=north,legend columns=-1},
            symbolic x coords={\rotatebox{90}{Telegram groups},\rotatebox{90}{Movie subtitles},\rotatebox{90}{Translated fiction}, \rotatebox{90}{Persian fiction},\rotatebox{90}{Blog posts},\rotatebox{90}{News comments},\rotatebox{90}{Product reviews},\rotatebox{90}{Online discussions}},
            xtick=data,
            nodes near coords,
            nodes near coords align={vertical},
            ymin=0,ymax=850,
            ylabel={Sentence count},
        ]
        \addplot[black,postaction={pattern=north east lines},fill=blue] table[x=direction,y=Development Data]{\mydata};
        \addplot[black,,postaction={pattern=north west lines},fill=red] table[x=direction,y=Test Data]{\mydata};
        \legend{Development Data, Test Data}
    \end{axis}
\end{tikzpicture}
}
\caption{Number of sentences from different genres in our evaluation development and test datasets.}
    \label{fig:genre}
\end{figure}
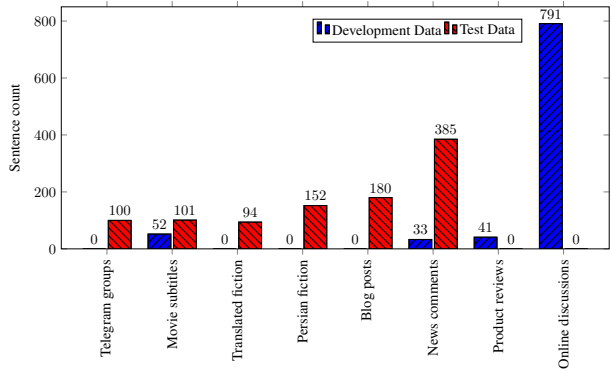
\section{Experiments}\label{sec:experiment}
In this section we describe our experiments and results. In addition to intrinsic evaluation on our evaluation datasets, we use machine translation as an extrinsic task for which the training data only contains colloquial text while the test data contains standard text.

\begin{figure*}[t!]
    \centering
    \includegraphics[width=.8\textwidth]{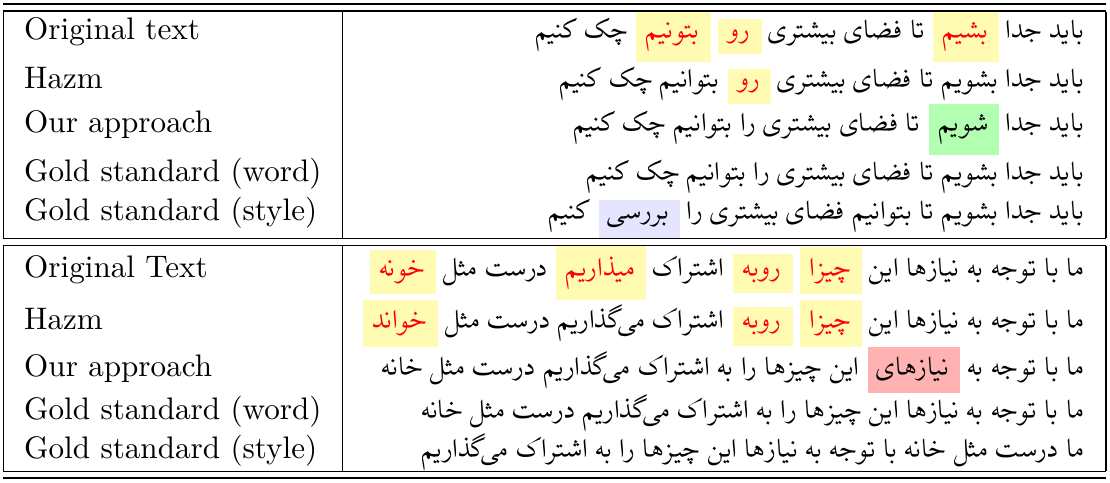}
    \caption{Examples of standardization outputs on our development data from our model and the baseline Hazm model. Wrong words are highlighted. We see that in the first example, although the output of our approach does not completely match with the surface word form, it is still a correct word (two correct forms of the same inflection). In the second example, we see that Hazm fails to do proper conversion, while our approach just adds an extra ``Ezafe'' suffix. }
    \label{fig:standard_example}
\end{figure*}

\subsection{Datasets and Tools}

\paragraph{Datasets:} We use the Wikipedia dump of Persian as standard text in addition to one million sentences from the Mizan corpus~\cite{kashefi2018mizan}. We randomly break  words and use this data as training data for standardization. Figure~\ref{fig:break_example} shows a few examples of converted text to colloquial. For machine translation evaluation, we use the TEP parallel data~\cite{pilevar2011tep} which is a collection of 600K translated movie subtitles in colloquial Persian. Following the splits by \newcite{khashabi20}, we use a small portion of the Mizan parallel corpus~\cite{kashefi2018mizan} (collection of classic fiction) as our development (1596 sentences) and test datasets (10000 sentences).

\paragraph{Tools and Baselines:} We use the Hazm library\footnote{\url{https://github.com/sobhe/hazm}} to normalize characters and tokenize texts. Hazm is also capable of converting colloquial text to standard text. This is done by manual rules. We use this tool as a strong baseline to compare with. We use SacreBLEU~\cite{post-2018-call} on the tokenized text for evaluation.

\paragraph{Translation Model:} We use a standard sequence-to-sequence transformer-based translation model~\cite{vaswani2017attention} with a six-layer BERT-based~\cite{devlin-etal-2019-bert} encoder-decoder architecture from HuggingFace~\cite{wolf2019huggingface} and Pytorch~\cite{paszke2019pytorch} with a shared SentencePiece~\cite{sentencepiece} vocabulary of size 40K.\footnote{Code from \url{https://github.com/rasoolims/ImageTranslate/tree/naacl21}} All input and output token embeddings are summed up with the language id embedding: we use ``$<$fa$>$'' for standard Persian, and ``$<$fab$>$'' for broken Persian. First tokens of every input and output sentence are shown by the language ID. We use greedy decoding since we find greedy decoding slightly more accurate for standardization than beam search. For machine translation, we use a similar pipeline. We pretrain the model on a tuple of three Wikipedia datasets for Persian, Arabic, and a sample of 6 million sentences from English using the MASS model~\cite{song2019mass} with a shared SentencePiece~\cite{sentencepiece} vocabulary of size 60K. The decoder for each language has a separate output layer in order to have language-specific output probabilities. We use a beam size of $4$ for machine translation experiments.

\subsection{Results}
\paragraph{Intrinsic evaluation:} Table~\ref{tab:evak_results} shows the BLEU scores of our model versus the rule-based Hazm tool. As we see in the Table, our model does a much better job in making the text closer to standard form. Figure~\ref{fig:standard_example} shows some examples of our conversions versus the rule-based conversion. It is worth noting that we observe some very few cases in which our model generates an irrelevant word, most likely due to the interference of the language model component in sequence-to-sequence models. Solving this issue is an interesting topic to pursue in future work.

\paragraph{Extrinsic evaluation:} Table~\ref{tab:mt_results} shows the BLEU scores of different models trained of different types of preprocessed and post-processed data. First, we observe that standardization helps in all cases. Second, using preprocessed data from our standardization model outperforms post-editing with  standardization. One important note about the low BLEU scores is that the TEP data~\cite{pilevar2011tep} consists of very short colloquial movie dialogues while the Mizan corpus~\cite{kashefi2018mizan} consists of sentences from classic fictions. This domain mismatch plus the size of the TEP data are the main causes of this low BLEU. However, out method still improves the BLEU score by a wide margin. In practice, previous work has used back-translation improve low-resource translation either in a non-iterative~\cite{edunov-etal-2018-understanding} or iterative ~\cite{hoang-etal-2018-iterative} manner.  Improvements from back-translation heavily depend on the quality of the initial model. We skip back-translation since it is not the focus of our work.

\begin{table}[t!]
    \centering  \setlength{\tabcolsep}{5pt}
    \begin{tabular}{l c c c c c }
    \hline  \hline 
     & \multicolumn{2}{c}{Development} & ~ &\multicolumn{2}{c}{Test} \\%\cline{2-5} \cline{7-10}
    & Word & Style & ~ &  Word & Style \\ \hline
    Original Data   &  43.9	& 38.8 & &	46.4 &		40.5 \\
    Hazm  &  	56.4 & 	49.0 & &  61.7	& 52.7 \\ % \hline
     Our approach & {\bf 61.7} & {\bf 53.9} &  & {\bf 62.8} &   {\bf 56.3 } \\ 
    %Our approach (beam=4) & {\bf 78.9}	& {\bf 58.9}	& {\bf 75.2}	& {\bf 51.5}	& &	79.8	& 61.1&	{\bf 79.1}	& {\bf 54.2} \\ 
    \hline  \hline 
    \end{tabular}
    \caption{Results of standardizing our Persian colloquial evaluation data via SacreBLEU~\protect\cite{post-2018-call}.}
    \label{tab:evak_results}
\end{table}
\begin{table}[t!]
    \centering  \setlength{\tabcolsep}{5pt}
    \begin{tabular}{ l l c c}
    \hline  \hline 
  Data &  Approach   & Dev. & Test \\ \hline
   Original Data   & No edit  & 4.4 & 2.8 \\ \hline
     \multirow{2}{*}{ Post-edit} & Hazm & 4.2 & 3.1 \\
     &  Our approach & 4.9 & 3.3  \\ \hline
   \multirow{2}{*}{Prepocess}& Hazm &  5.0 & 3.0 \\ 
    & Our approach & {\bf 5.8} & {\bf 3.6}  \\ \hline  \hline 
    \end{tabular}
    \caption{Machine translation results trained on TEP~\protect\cite{pilevar2011tep} (movie subtitles in colloquial Persian), and evaluated on the Mizan corpus~\protect\cite{kashefi2018mizan} (standard Persian) via SacreBLEU~\protect\cite{post-2018-call}.}
    \label{tab:mt_results}
\end{table}

\section{Conclusion}\label{sec:conclude}
We have described an algorithm for standardizing Persian text. We show that by creating artificial training data, we can leverage commonly known patterns of breaking standard forms to colloquial, and learn an accurate standardization model. We believe that this work is just a beginning to this line of research. Future work should investigate more sophisticated methods for standardization.

\section*{Acknowledgment}
We would like to deeply thank Omid Tabibzadeh for sharing his knowledge with us and constructive discussions.

\bibliography{refs}

\begin{thebibliography}{35}
\expandafter\ifx\csname natexlab\endcsname\relax\def\natexlab#1{#1}\fi

\bibitem[{Armin and Shamsfard(2011)}]{armin}
Nadieh Armin and Mehrnoush Shamsfard. 2011.
\newblock Converting {Persian} colloquium text to formal by n-grams.
\newblock In \emph{Computer Society of Iran}.

\bibitem[{Bahdanau et~al.(2015)Bahdanau, Cho, and
  Bengio}]{Bahdanau2015NeuralMT}
Dzmitry Bahdanau, Kyunghyun Cho, and Yoshua Bengio. 2015.
\newblock Neural machine translation by jointly learning to align and
  translate.
\newblock \emph{CoRR}, abs/1409.0473.

\bibitem[{Bakhshizadeh~Gashti and Tabibzadeh(2019)}]{tabib_journal}
Yousef Bakhshizadeh~Gashti and Omid Tabibzadeh. 2019.
\newblock Colloquial forms and {Persian} lexicons.
\newblock \emph{Persian Language and Iranian Dialects}, 2(6).

\bibitem[{Bijankhan et~al.(2011)Bijankhan, Sheykhzadegan, Bahrani, and
  Ghayoomi}]{bijankhan2011lessons}
Mahmood Bijankhan, Javad Sheykhzadegan, Mohammad Bahrani, and Masood Ghayoomi.
  2011.
\newblock Lessons from building a {Persian} written corpus: {Peykare}.
\newblock \emph{Language resources and evaluation}, 45(2):143--164.

\bibitem[{Boyle(1952)}]{boyle1952notes}
John~Andrew Boyle. 1952.
\newblock Notes on the colloquial language of {Persia} as recorded in certain
  recent writings.
\newblock \emph{Bulletin of the School of Oriental and African Studies},
  14(3):451--462.

\bibitem[{Cho et~al.(2014)Cho, van Merri{\"e}nboer, Gulcehre, Bahdanau,
  Bougares, Schwenk, and Bengio}]{cho-etal-2014-learning}
Kyunghyun Cho, Bart van Merri{\"e}nboer, Caglar Gulcehre, Dzmitry Bahdanau,
  Fethi Bougares, Holger Schwenk, and Yoshua Bengio. 2014.
\newblock \href {https://doi.org/10.3115/v1/D14-1179} {Learning phrase
  representations using {RNN} encoder{--}decoder for statistical machine
  translation}.
\newblock In \emph{Proceedings of the 2014 Conference on Empirical Methods in
  Natural Language Processing ({EMNLP})}, pages 1724--1734, Doha, Qatar.
  Association for Computational Linguistics.

\bibitem[{Devlin et~al.(2019)Devlin, Chang, Lee, and
  Toutanova}]{devlin-etal-2019-bert}
Jacob Devlin, Ming-Wei Chang, Kenton Lee, and Kristina Toutanova. 2019.
\newblock \href {https://doi.org/10.18653/v1/N19-1423} {{BERT}: Pre-training of
  deep bidirectional transformers for language understanding}.
\newblock In \emph{Proceedings of the 2019 Conference of the North {A}merican
  Chapter of the Association for Computational Linguistics: Human Language
  Technologies, Volume 1 (Long and Short Papers)}, pages 4171--4186,
  Minneapolis, Minnesota. Association for Computational Linguistics.

\bibitem[{Edunov et~al.(2018)Edunov, Ott, Auli, and
  Grangier}]{edunov-etal-2018-understanding}
Sergey Edunov, Myle Ott, Michael Auli, and David Grangier. 2018.
\newblock \href {https://doi.org/10.18653/v1/D18-1045} {Understanding
  back-translation at scale}.
\newblock In \emph{Proceedings of the 2018 Conference on Empirical Methods in
  Natural Language Processing}, pages 489--500, Brussels, Belgium. Association
  for Computational Linguistics.

\bibitem[{Feely et~al.(2014)Feely, Manshadi, Frederking, and
  Levin}]{feely2014cmu}
Weston Feely, Mehdi Manshadi, Robert~E. Frederking, and Lori~S. Levin. 2014.
\newblock The {CMU METAL Farsi NLP} approach.
\newblock In \emph{LREC}, pages 4052--4055.

\bibitem[{Hoang et~al.(2018)Hoang, Koehn, Haffari, and
  Cohn}]{hoang-etal-2018-iterative}
Vu~Cong~Duy Hoang, Philipp Koehn, Gholamreza Haffari, and Trevor Cohn. 2018.
\newblock \href {https://doi.org/10.18653/v1/W18-2703} {Iterative
  back-translation for neural machine translation}.
\newblock In \emph{Proceedings of the 2nd Workshop on Neural Machine
  Translation and Generation}, pages 18--24, Melbourne, Australia. Association
  for Computational Linguistics.

\bibitem[{Jeremi{\'a}s(1984)}]{jeremias1984diglossia}
{\'E}va~M Jeremi{\'a}s. 1984.
\newblock Diglossia in persian.
\newblock \emph{Acta Linguistica Academiae Scientiarum Hungaricae},
  34(3/4):271--287.

\bibitem[{Kashefi(2018)}]{kashefi2018mizan}
Omid Kashefi. 2018.
\newblock Mizan: a large {Persian-English} parallel corpus.
\newblock \emph{arXiv preprint arXiv:1801.02107}.

\bibitem[{Khashabi et~al.(2020)Khashabi, Cohan, Shakeri, Hosseini, Pezeshkpour,
  Alikhani, Aminnaseri, Bitaab, Brahman, Ghazarian, Gheini, Kabiri, Mahabadi,
  Memarrast, Mosallanezhad, Noury, Raji, Rasooli, Sadeghi, Azer, Samghabadi,
  Shafaei, Sheybani, Tazarv, and Yaghoobzadeh}]{khashabi20}
Daniel Khashabi, Arman Cohan, Siamak Shakeri, Pedram Hosseini, Pouya
  Pezeshkpour, Malihe Alikhani, Moin Aminnaseri, Marzieh Bitaab, Faeze Brahman,
  Sarik Ghazarian, Mozhdeh Gheini, Arman Kabiri, Rabeeh~Karimi Mahabadi, Omid
  Memarrast, Ahmadreza Mosallanezhad, Erfan Noury, Shahab Raji, Mohammad~Sadegh
  Rasooli, Sepideh Sadeghi, Erfan~Sadeqi Azer, Niloofar~Safi Samghabadi, Mahsa
  Shafaei, Saber Sheybani, Ali Tazarv, and Yadollah Yaghoobzadeh. 2020.
\newblock {\textsc{ParsiNLU}}: A suite of language understanding challenges for
  {Persian}.
\newblock \emph{arXiv preprint}.

\bibitem[{Kudo and Richardson(2018)}]{sentencepiece}
Taku Kudo and John Richardson. 2018.
\newblock \href {https://doi.org/10.18653/v1/D18-2012} {{S}entence{P}iece: A
  simple and language independent subword tokenizer and detokenizer for neural
  text processing}.
\newblock In \emph{Proceedings of the 2018 Conference on Empirical Methods in
  Natural Language Processing: System Demonstrations}, pages 66--71, Brussels,
  Belgium. Association for Computational Linguistics.

\bibitem[{Mahmoodi-Bakhtiari(2018)}]{bakh2018}
Behrooz Mahmoodi-Bakhtiari. 2018.
\newblock \href {https://doi.org/https://doi.org/10.1515/9783110455793} {Spoken
  vs. written {Persian}: Is {Persian} diglossic?}
\newblock In Alireza Korangy and Corey Miller, editors, \emph{Trends in Iranian
  and Persian Linguistics}, chapter~10, pages 183--212. De Gruyter Mouton.

\bibitem[{Mirzaei and Moloodi(2016)}]{mirzaei2016persian}
Azadeh Mirzaei and Amirsaeid Moloodi. 2016.
\newblock {Persian} proposition bank.
\newblock In \emph{Proceedings of the Tenth International Conference on
  Language Resources and Evaluation (LREC'16)}, pages 3828--3835.

\bibitem[{Mirzaei and Safari(2018)}]{mirzaei2018persian}
Azadeh Mirzaei and Pegah Safari. 2018.
\newblock {Persian} discourse treebank and coreference corpus.
\newblock In \emph{Proceedings of the Eleventh International Conference on
  Language Resources and Evaluation (LREC 2018)}.

\bibitem[{Mirzaei et~al.(2020)Mirzaei, Sedghi, and
  Safari}]{mirzaei2020semantic}
Azadeh Mirzaei, Fatemeh Sedghi, and Pegah Safari. 2020.
\newblock Semantic role labeling system for persian language.
\newblock \emph{ACM Transactions on Asian and Low-Resource Language Information
  Processing (TALLIP)}, 19(3):1--12.

\bibitem[{Nourian et~al.(2015)Nourian, Rasooli, Imany, and
  Faili}]{nourian2015importance}
Alireza Nourian, Mohammad~Sadegh Rasooli, Mohsen Imany, and Heshaam Faili.
  2015.
\newblock On the importance of {Ezafe} construction in {Persian} parsing.
\newblock In \emph{Proceedings of the 53rd Annual Meeting of the Association
  for Computational Linguistics and the 7th International Joint Conference on
  Natural Language Processing (Volume 2: Short Papers)}, pages 877--882.

\bibitem[{Paszke et~al.(2019)Paszke, Gross, Massa, Lerer, Bradbury, Chanan,
  Killeen, Lin, Gimelshein, Antiga et~al.}]{paszke2019pytorch}
Adam Paszke, Sam Gross, Francisco Massa, Adam Lerer, James Bradbury, Gregory
  Chanan, Trevor Killeen, Zeming Lin, Natalia Gimelshein, Luca Antiga, et~al.
  2019.
\newblock Pytorch: An imperative style, high-performance deep learning library.
\newblock In \emph{Advances in neural information processing systems}, pages
  8026--8037.

\bibitem[{Pilevar et~al.(2011)Pilevar, Faili, and Pilevar}]{pilevar2011tep}
Mohammad~Taher Pilevar, Heshaam Faili, and Abdol~Hamid Pilevar. 2011.
\newblock {TEP}: {Tehran} {English-Persian} parallel corpus.
\newblock In \emph{International Conference on Intelligent Text Processing and
  Computational Linguistics}, pages 68--79. Springer.

\bibitem[{Poostchi et~al.(2018)Poostchi, Borzeshi, and
  Piccardi}]{poostchi2018bilstm}
Hanieh Poostchi, Ehsan~Zare Borzeshi, and Massimo Piccardi. 2018.
\newblock {BiLSTM-CRF} for {Persian} named-entity recognition.
  {ArmanPersoNERCorpus}: The first entity-annotated {Persian} dataset.
\newblock In \emph{Proceedings of the Eleventh International Conference on
  Language Resources and Evaluation (LREC 2018)}.

\bibitem[{Post(2018)}]{post-2018-call}
Matt Post. 2018.
\newblock \href {https://doi.org/10.18653/v1/W18-6319} {A call for clarity in
  reporting {BLEU} scores}.
\newblock In \emph{Proceedings of the Third Conference on Machine Translation:
  Research Papers}, pages 186--191, Brussels, Belgium. Association for
  Computational Linguistics.

\bibitem[{Rasooli et~al.(2013)Rasooli, Kouhestani, and
  Moloodi}]{rasooli2013development}
Mohammad~Sadegh Rasooli, Manouchehr Kouhestani, and Amirsaeid Moloodi. 2013.
\newblock Development of a {Persian} syntactic dependency treebank.
\newblock In \emph{Proceedings of the 2013 Conference of the North American
  Chapter of the Association for Computational Linguistics: Human Language
  Technologies}, pages 306--314.

\bibitem[{Sennrich et~al.(2016)Sennrich, Haddow, and
  Birch}]{sennrich-etal-2016-improving}
Rico Sennrich, Barry Haddow, and Alexandra Birch. 2016.
\newblock \href {https://doi.org/10.18653/v1/P16-1009} {Improving neural
  machine translation models with monolingual data}.
\newblock In \emph{Proceedings of the 54th Annual Meeting of the Association
  for Computational Linguistics (Volume 1: Long Papers)}, pages 86--96, Berlin,
  Germany. Association for Computational Linguistics.

\bibitem[{Seraji(2015)}]{seraji2015morphosyntactic}
Mojgan Seraji. 2015.
\newblock \emph{Morphosyntactic corpora and tools for {Persian}}.
\newblock Ph.D. thesis, Acta Universitatis Upsaliensis.

\bibitem[{Seraji et~al.(2016)Seraji, Ginter, and Nivre}]{seraji2016universal}
Mojgan Seraji, Filip Ginter, and Joakim Nivre. 2016.
\newblock Universal dependencies for {Persian}.
\newblock In \emph{Proceedings of the Tenth International Conference on
  Language Resources and Evaluation (LREC'16)}, pages 2361--2365.

\bibitem[{Shamsfard(2011)}]{shamsfard2011challenges}
Mehrnoush Shamsfard. 2011.
\newblock Challenges and open problems in {Persian} text processing.
\newblock In \emph{Proceedings of LTC}.

\bibitem[{Solhju(2019)}]{solhju}
Ali Solhju. 2019.
\newblock \emph{How To Write Spoken Short Forms In Persian Dialogues}.
\newblock Markaz, Tehran, Iran.

\bibitem[{Song et~al.(2019)Song, Tan, Qin, Lu, and Liu}]{song2019mass}
Kaitao Song, Xu~Tan, Tao Qin, Jianfeng Lu, and Tie-Yan Liu. 2019.
\newblock {MASS}: Masked sequence to sequence pre-training for language
  generation.
\newblock \emph{arXiv cs.CL 1905.02450}.

\bibitem[{Tabibzadeh(2020)}]{tabib_book}
Omid Tabibzadeh. 2020.
\newblock \emph{Orthography of Colloquial Persian: Based on Works of Fiction
  and Drama Spanning a Century (1918-2018)}.
\newblock Institute for Humanities and Cultural Studies, Tehran, Iran.

\bibitem[{Taher et~al.(2020)Taher, Hoseini, and Shamsfard}]{taher2020beheshti}
Ehsan Taher, Seyed~Abbas Hoseini, and Mehrnoush Shamsfard. 2020.
\newblock {Beheshti-NER}: {Persian} named entity recognition using {BERT}.
\newblock \emph{arXiv preprint arXiv:2003.08875}.

\bibitem[{Vaswani et~al.(2017)Vaswani, Shazeer, Parmar, Uszkoreit, Jones,
  Gomez, Kaiser, and Polosukhin}]{vaswani2017attention}
Ashish Vaswani, Noam Shazeer, Niki Parmar, Jakob Uszkoreit, Llion Jones,
  Aidan~N Gomez, {\L}ukasz Kaiser, and Illia Polosukhin. 2017.
\newblock Attention is all you need.
\newblock In \emph{Advances in neural information processing systems}, pages
  5998--6008.

\bibitem[{Windfuhr(2009)}]{windfuhr2009iranian}
Gernot Windfuhr. 2009.
\newblock \emph{The Iranian Languages}.
\newblock Psychology Press.

\bibitem[{Wolf et~al.(2019)Wolf, Debut, Sanh, Chaumond, Delangue, Moi, Cistac,
  Rault, Louf, Funtowicz et~al.}]{wolf2019huggingface}
Thomas Wolf, Lysandre Debut, Victor Sanh, Julien Chaumond, Clement Delangue,
  Anthony Moi, Pierric Cistac, Tim Rault, R{\'e}mi Louf, Morgan Funtowicz,
  et~al. 2019.
\newblock Huggingface's transformers: State-of-the-art natural language
  processing.
\newblock \emph{ArXiv}, pages arXiv--1910.

\end{thebibliography}
\bibliographystyle{acl_natbib}
\end{document}